\documentclass[10pt,twocolumn,letterpaper]{article}

\usepackage[pagenumbers]{wacv} 

\usepackage[T1]{fontenc}
\usepackage{graphicx}
\usepackage{amsmath}
\usepackage{amssymb}
\usepackage{booktabs}

\usepackage{multirow}
\usepackage{ulem}
\usepackage{subcaption}

\usepackage[pagebackref,breaklinks,colorlinks]{hyperref}

\usepackage[capitalize]{cleveref}
\crefname{section}{Sec.}{Secs.}
\Crefname{section}{Section}{Sections}
\Crefname{table}{Table}{Tables}
\crefname{table}{Tab.}{Tabs.}

\def\wacvPaperID{5} 
\def\confName{WACV}
\def\confYear{2025}

\begin{document}

\title{RAPTOR: Refined Approach for Product Table Object Recognition}

\author{Eliott THOMAS\\
Yooz, France\\
{\tt\small eliott.thomas@getyooz.com}
\and
Mickael COUSTATY\\
La Rochelle Université, France\\
{\tt\small mickael.coustaty@univ-lr.fr}
\and
Aurélie JOSEPH\\
Yooz, France\\
{\tt\small aurelie.joseph@getyooz.com}
\and
Gaspar DELOIN\\
Yooz, France\\
{\tt\small gaspar.deloin@getyooz.com}
\and
Elodie CAREL\\
Yooz, France\\
{\tt\small elodie.carel@getyooz.com}
\and
Vincent POULAIN D'ANDECY\\
Yooz, France\\
{\tt\small Vincent.PoulaindAndecy@getyooz.com}
\and
Jean-Marc OGIER\\
La Rochelle Université, France\\
{\tt\small jean-marc.ogier@univ-lr.fr}
}

\maketitle

\begin{abstract}

Extracting tables from documents is a critical task across various industries, especially on business documents like invoices and reports. Existing systems based on DEtection TRansformer (DETR) such as TAble TRansformer (TATR), offer solutions for Table Detection (TD) and Table Structure Recognition (TSR) but face challenges with diverse table formats and common errors like incorrect area detection and overlapping columns. This research introduces RAPTOR, a modular post-processing system designed to enhance state-of-the-art models for improved table extraction, particularly for product tables. RAPTOR, addresses recurrent TD and TSR issues, improving both precision and structural predictions. For TD, we use DETR (trained on ICDAR 2019) and TATR (trained on PubTables-1M and FinTabNet), while TSR only relies on TATR. A Genetic Algorithm is incorporated to optimize RAPTOR’s module parameters, using a private dataset of product tables to align with industrial needs. We evaluate our method on two private datasets of product tables, the public DOCILE dataset (which contains tables similar to our target product tables), and the ICDAR 2013 and ICDAR 2019 datasets. The results demonstrate that while our approach excels at product tables, it also maintains reasonable performance across diverse table formats. An ablation study further validates the contribution of each module in our system.


\end{abstract}

\section{Introduction}
\label{sec:intro}

Efficient extraction of tabular data is crucial across various industries, particularly in business applications where tables are commonly used to present structured information such as sales figures and financial performance. Two key processes are involved in extracting information from these tables: TD, which identifies table boundaries, and TSR, which defines the internal layout of rows, columns and headers. Accurately detecting and extracting table content remains a challenging task due to the diversity of table formats and the limited availability of annotated datasets, especially for business tables, which often contain proprietary data.

Among the various types of tables, product tables are specialized formats that include data such as product names, descriptions, prices, and supplier information. These tables are essential for inventory management, order processing, and strategic decision-making. In this work, we specifically focus on simple product tables, which do not include complicated structures such as spanning cells \cite{Chi2019ComplicatedTS}. 

Given the scarcity of annotated business documents for privacy reasons, fine-tuning existing models is challenging. Thus, we utilize pretrained models on public datasets: DETR is pretrained on ICDAR 2019 for TD, and TATR is pretrained on Pubtables-1M and FinTabNet for both TD and TSR \cite{Carion2020EndtoEndOD,Gao2019ICDAR2C,Smock2021PubTables1MTC,Zheng2020GlobalTE}. The goal of our modular method is to provide an alternative to fine-tuning when only a limited amount of training data is available. To achieve this, we apply a Genetic Algorithm to learn the parameters of our modules using a very limited amount of private data. This modular strategy allows us to leverage the strengths of different models while tailoring our approach to specific challenges in table extraction.

We evaluate RAPTOR on five datasets, including two private and three public ones: ICDAR 2013, ICDAR 2019, and a curated subset of the DocILE dataset \cite{Gbel2013ICDAR2T,Gao2019ICDAR2C,vSimsa2023DocILEBF}, which we release publicly for table extraction tasks. Our results show that RAPTOR improves precision and resolves common extraction issues, enhancing baseline model performance across datasets. We identify the most common types of errors in business tables, demonstrating RAPTOR’s adaptability to various table extraction tools. Our contributions can be summarized as follow:
\begin{itemize}
    \renewcommand{\labelitemi}{ \textbullet} 
    \item Identification of the most common types of errors in business tables
    \item Introduction of a modular system, RAPTOR, to help mitigate these errors
    \item Release of a subset of the DocILE dataset for TD and TSR tasks
\end{itemize}

\begin{table*}[h!]
\centering

\begin{tabular}{l c c | l c c}
\toprule
\multicolumn{2}{c}{\textbf{Business (Private)}} & \multicolumn{1}{c|}{} & \multicolumn{2}{c}{\textbf{ICDAR 2019 (Public)}} & \multicolumn{1}{c}{\textbf{}} \\ 
\midrule
\textbf{Error Name}   & \textbf{Freq [\%]} & \textbf{Cat} & \textbf{Error Name}   & \textbf{Freq [\%]} & \textbf{Cat} \\ \midrule
Noise detected below & 22               & TD & Implicit header       & 32               & TSR \\ 
Wrong table detected  & 18               & TD & Spanning cells        & 26               & TSR \\ 
Missing elements      & 13               & TSR & Missing elements      & 12               & TSR \\ 
Noise detected above  & 11               & TD & Wrong header detected & 11               & TSR \\ 
False positive        & 8                & TD & Segmentation error    & 5                & TSR \\ 
Wrong header detected & 6                & TSR & Sub-tables            & 4                & TSR \\ 
Segmentation error    & 4                & TSR & Wrong table detected  & 2                & TD \\ 
Fused Lines           & 3                & TSR & Splitted line         & 2                & TSR \\ 
\bottomrule
\end{tabular}
\caption{Most common errors in Business and ICDAR 2019 datasets, with their frequency among total errors.}
\label{error_table}
\end{table*}

\section{Related Work}
\label{sec:related-work}

\subsection{Table Detection (TD)}

Table extraction from documents has been a longstanding research topic, with early surveys like Zanibbi et al. \cite{Zanibbi2004ASO} highlighting its significance. While image-based models like Faster R-CNN \cite{Ren2015FasterRT} have been employed, they often underperform due to the unique characteristics and variability of tables \cite{Smock2021PubTables1MTC}. Alexiou et al. \cite{Alexiou2023AnEO} provide a succinct review of traditional table detection methods, categorizing them into heuristic, deep learning, and hybrid approaches. Rujiao et al. introduce the WTW dataset \cite{Long2021ParsingTS} to address realistic scenarios, such as images of documents taken in real-world conditions—like photos of papers, screenshots, or scanned documents with various distortions, including perspective warping and blurring. However, our approach assumes access to deskewed scanned documents paired with OCR technologies. We use ABBYY OCR for our private data and DocTR \cite{doctr2021} for public datasets.

We leverage DETR \cite{Carion2020EndtoEndOD}, which merges a CNN backbone for feature extraction with a transformer architecture to model global context. This innovative architecture mitigates the need for post-processing steps like Non-Maximum Suppression (NMS) by employing a bipartite matching loss during training. For our table detection tasks, we utilize DETR-19, trained specifically on the ICDAR 2019 dataset. Alongside it we used Table Transformer (TATR), an extension of DETR fine-tuned on table-specific datasets like PubTables-1M \cite{Smock2021PubTables1MTC} and FinTabNet \cite{Zheng2020GlobalTE}. This combination enhances our ability to detect tables across diverse document layouts.

\subsection{Table Structure Recognition (TSR)}

Graph-based methods were once promising for table extraction due to their ability to model relationships between table elements. Techniques by Riba et al. \cite{Riba2019TableDI}, Zheng et al. \cite{Zheng2020GlobalTE}, and Li et al. \cite{Li2022TableSR} used Graph Neural Networks (GNNs) to capture the structure of tables. However, they no longer offer state-of-the-art (SOTA) results, having been surpassed by deep learning models like DETR and TATR, which excel at directly predicting table boundaries and structures. TableNet provides an end-to-end solution for table detection and structure recognition by highlighting text zones with different colors based on their type, as shown in Paliwal et al. \cite{Paliwal2019TableNetDL}. Similarly, Prasad et al. \cite{Prasad2020CascadeTabNetAA} use the image-based CascadeTabNet, which employs iterative transfer learning and data augmentation to enhance performance. Rashid et al. \cite{Rashid2017TableRI} introduce a hybrid method (image and text) that predicts whether segments of a document are inside or outside of tables, enhancing accuracy by averaging predictions with their neighbors.

For table structure recognition, we used TATR \cite{Smock2021PubTables1MTC} as well. TATR for TSR is designed to capture the structural relationships within tables, such as rows and columns, making it highly effective for TSR tasks. By using TATR in both TD and TSR, we ensure a comprehensive approach that leverages its strengths to effectively tackle the challenges of table extraction in real-world scenarios.

\subsection{Challenges in Business Table Extraction}

Despite significant advancements in table extraction, a notable gap exists in research focused on business tables, particularly product tables. Business tables, as seen in datasets like DocILE\cite{vSimsa2023DocILEBF}(Section \ref{sec:datasets}), encompass a variety of data types, such as product details or invoices. They often exhibit considerable noise, with content added arbitrarily and inconsistent spacing between lines, complicating the extraction process. In contrast, datasets like FinTabNet \cite{Zheng2020GlobalTE} specifically target financial documents, which are typically clean and well-structured, focusing on numerical metrics and financial reporting. 

\begin{figure*}[t]
  \centering
   \includegraphics[width=1.0\linewidth]{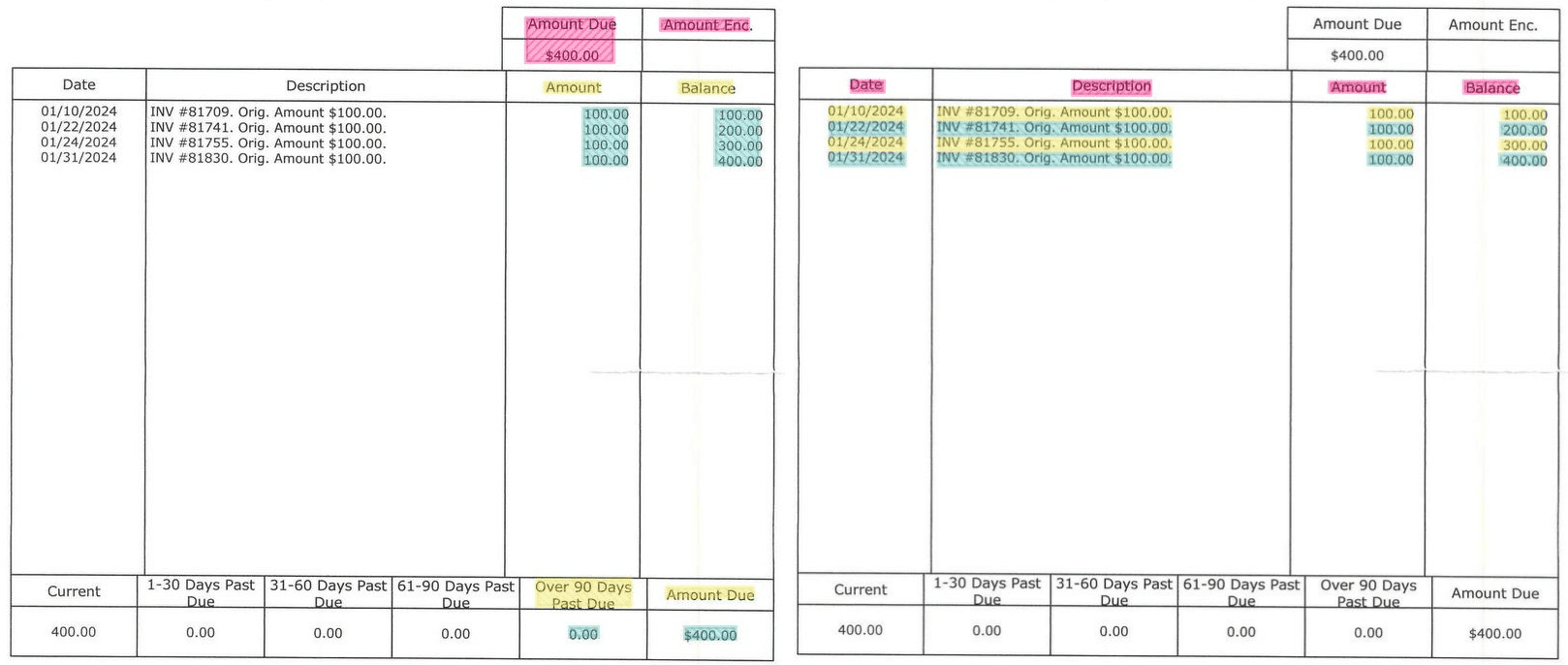}

   \caption{Baseline prediction VS Ground-Truth on anonymised private data. Common errors (on the left) generally include mistakes in line fusion and over detected cells that can be seen as noise for the structure recognition step.}
   \label{fig:example}
\end{figure*}

In Figure \ref{fig:example}, we illustrate common errors made by the baselines, including noise above and below the GT table and poorly fused lines. Our analysis of common errors in Table \ref{error_table} on the Business (private) dataset shows a high prevalence of TD errors, often due to business-specific elements resembling tables, like tax indications. To address these issues, we propose a post-processing method with corrective modules targeting frequent TD and TSR errors. These modules, optimized using a Genetic Algorithm (GA)\cite{holland1992adaptation}, improve accuracy after the initial model predictions. This approach contrasts with Le Vine et al.'s method \cite{Vine2019ExtractingTF}, which uses GA and Conditional Generative Adversarial Networks (cGAN) \cite{Isola2016ImagetoImageTW} to generate table structures directly. We anticipate that our method will generalize well to other public business datasets, such as DocILE, due to their shared domain. In contrast, errors from the ICDAR 2019 dataset highlight the varied challenges across different document types.

\section{Methodology}
\label{sec:methodology}

\begin{figure*}[t]
  \centering
   \includegraphics[width=0.99\linewidth]{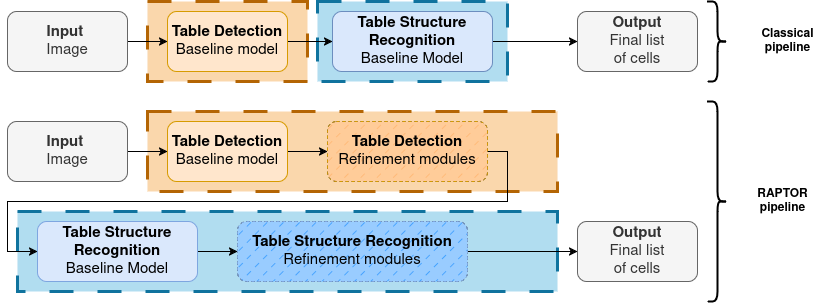}

   \caption{Classical Table Extraction pipeline vs RAPTOR pipeline.}
   \label{fig:pipeline}
\end{figure*}

Our pipeline (see Figure \ref{fig:pipeline}), integrates modular components to enhance both TD and TSR. For TD, we employ an ensemble approach that combines predictions from the DETR and TATR models to identify table regions within documents. These initial detections are further refined through dedicated TD modules. Once the table region is confirmed, the TSR task is carried out using the TATR model, with additional refinement modules improving table structure recognition. This flexible, modular design allows the pipeline to adapt to different baseline models and dataset-specific needs.

\subsection{Table Detection refinement}

The TD task aims to extract the overall table area. However, when this area is processed by the TSR model, it assumes that all content within the detected boundary is relevant. This is often valid for ICDAR datasets, but business tables frequently contain additional elements, such as empty spaces and noise. Thus, TD refinement adjusts the detected boundaries to better fit the smallest bounding box containing only the essential content for TSR.

\subsubsection{Table Chooser}
\label{sec:bbox-chooser}

RAPTOR utilizes a Table Chooser module to refine candidate regions identified by the TD baseline (the ensemble of DETR and TATR models). Given the monotable context, where only one table needs to be selected, we cannot rely solely on the confidence scores of the predictions. Detectors may erroneously predict table-like objects with higher confidence than the actual table. Our Table Chooser identifies the bounding box that most likely represents a table in the document image.

Initially, candidate regions with confidence scores above a threshold $\alpha$ are considered. The selection process employs a multi-feature evaluation strategy, incorporating prediction confidence, geometric properties (area, width, and height), and textual content analysis. The textual analysis uses the Levenshtein distance~\cite{Levenshtein1965BinaryCC} to compare candidates against a dictionary of potential table headers generated by ChatGPT 3.5 from the provided examples.

The module also focuses on header location within the bounding box, with a greater concentration towards the top indicating a higher likelihood of being a table. It assigns scores to each bounding box and selects the highest-scoring one for further processing. This module is crucial for resolving the "Wrong table detected" issue (noted in Table \ref{error_table}), which frequently arises in our datasets due to single tables being surrounded by table-like entities. The detailed dictionary used for textual analysis is available in the \href{https://anonymous.4open.science/r/RAPTOR_data-CF2C/README.md}{Data Repository}.

\subsubsection{Header Refiner}
\label{sec:header-refiner}

The Header Refiner module enhances the bounding box selected by the Table Chooser to ensure the top portion accurately represents the table header. This iterative process checks for content above the header, including content alignment (left or right), common header keywords, and stopwords that should not appear in headers. If any checks raise concerns about the top area, the module crops the bounding box until a valid header is identified. This refinement ensures that subsequent processing in the RAPTOR pipeline focuses on accurate table header recognition, effectively addressing the "Noise detected above" issue noted in Table \ref{error_table}.

\subsubsection{Oversegmentation management}
\label{sec:pruner}

This module addresses the "Noise detected below" issue from Table \ref{error_table}. RAPTOR introduces the oversegmentation module which uses DBSCAN clustering~\cite{Ester1996ADA} to group words by their y-coordinates, estimating horizontal lines. X-axis and XY-axis clustering techniques are applied to detect column position outliers and overall table structure. RAPTOR analyzes the sequence of clusters to identify the most frequent pattern, representing the "core table line". Starting from the bottom of the table, each line receives a suspiciousness score based on the number of outliers and its deviation from the core pattern (using Levenshtein distance~\cite{Levenshtein1965BinaryCC}). Suspicious lines are removed, and the bounding box is adjusted upward until a non-suspicious core line is reached.

\subsection{Table Structure Recognition refinement}

In the TSR task, our goal is to accurately interpret content within the detected table area. However, the complexity of business tables, often characterized by noise and inconsistent formatting, can lead to inaccuracies when directly transferring data from the TD model. To improve the quality of table structures, we implement refinement techniques that resolve common issues such as overlapping columns, unfused lines, and extraneous lines.

\subsubsection{Overlapping columns}
\label{sec:overlap}

The Overlapping Columns module helps mitigate the "Segmentation error" issue from Table \ref{error_table}. RAPTOR detects overlapping columns using the Intersection-over-Union (IoU) metric \cite{Everingham2010ThePV}. When columns with an IoU greater than a specified threshold \( \theta_{\text{IoU}} \) are detected, RAPTOR compares the number of header words (identified by the TATR model) in each column and selects the one with the most headers words as the primary column.

\subsubsection{Unfuse Lines}
\label{sec:unfuser}
In tables, rows may sometimes include multiline entries, such as lengthy comments or descriptions. The Unfuse Lines module in RAPTOR addresses the "Fused Lines" issue identified in Table \ref{error_table}. This module focuses specifically on cases where cells within a row contain multiple amounts. It detects and separates these merged rows into distinct entities. Additionally, the module employs spatial coordinates to split other cells within the same row.

\subsubsection{Remove Extra Lines}
\label{sec:remove-line}
This approach helps mitigate the "Segmentation error" and "Noise detected below" issues identified in Table \ref{error_table}. It eliminates lines from the bottom of the table that do not conform to expected alignment or data type similarities. Additionally, we check whether the last lines are too far from the rest of the table by measuring the space between lines. If the gap exceeds $\beta$ times the largest height among the predicted lines, those lines are flagged and removed.

\subsection{Genetic Algorithm Process}
\label{sec:GA}
Genetic Algorithms (GAs) are optimization techniques inspired by natural selection \cite{holland1992adaptation}. In our methodology, we applied a GA to optimize RAPTOR's parameters using a refinement dataset similar to our target data. We initialized a population of candidate solutions, each representing a set of module parameters. The GA evolved this population over several generations, optimizing based on the F1 score from GRITS-CON \cite{Smock2022GriTSGT}. Key evolutionary mechanisms included elitism, crossover, and mutation, with the best-performing individual from the final generation defining the new parameters for RAPTOR.

\section{Evaluation protocol}
\label{sec:datasets}

\subsection{Metrics}

In our evaluation of RAPTOR's performance, we used various metrics. For TD, we employed Purity and Completeness \cite{Gbel2013ICDAR2T}: Purity measures the proportion of predicted tables whose elements are fully contained within ground truth (GT) tables, while Completeness assesses the predicted tables that encompass all GT elements. Additionally, we applied the Generalized Intersection over Union (GIoU) \cite{Rezatofighi2019GeneralizedIO}, which extends traditional IoU by considering predictions without overlap, ranging from -1 to 1. For TSR, we selected GRITS, which overcomes limitations of previous metrics, as noted in \cite{Smock2022GriTSGT}, specifically GRITS-CON to compare cell content between GT and predicted tables.

\subsection{Inference Datasets}

\begin{figure*}
  \centering
  \begin{subfigure}{0.32\linewidth}
    \includegraphics[width=\linewidth]{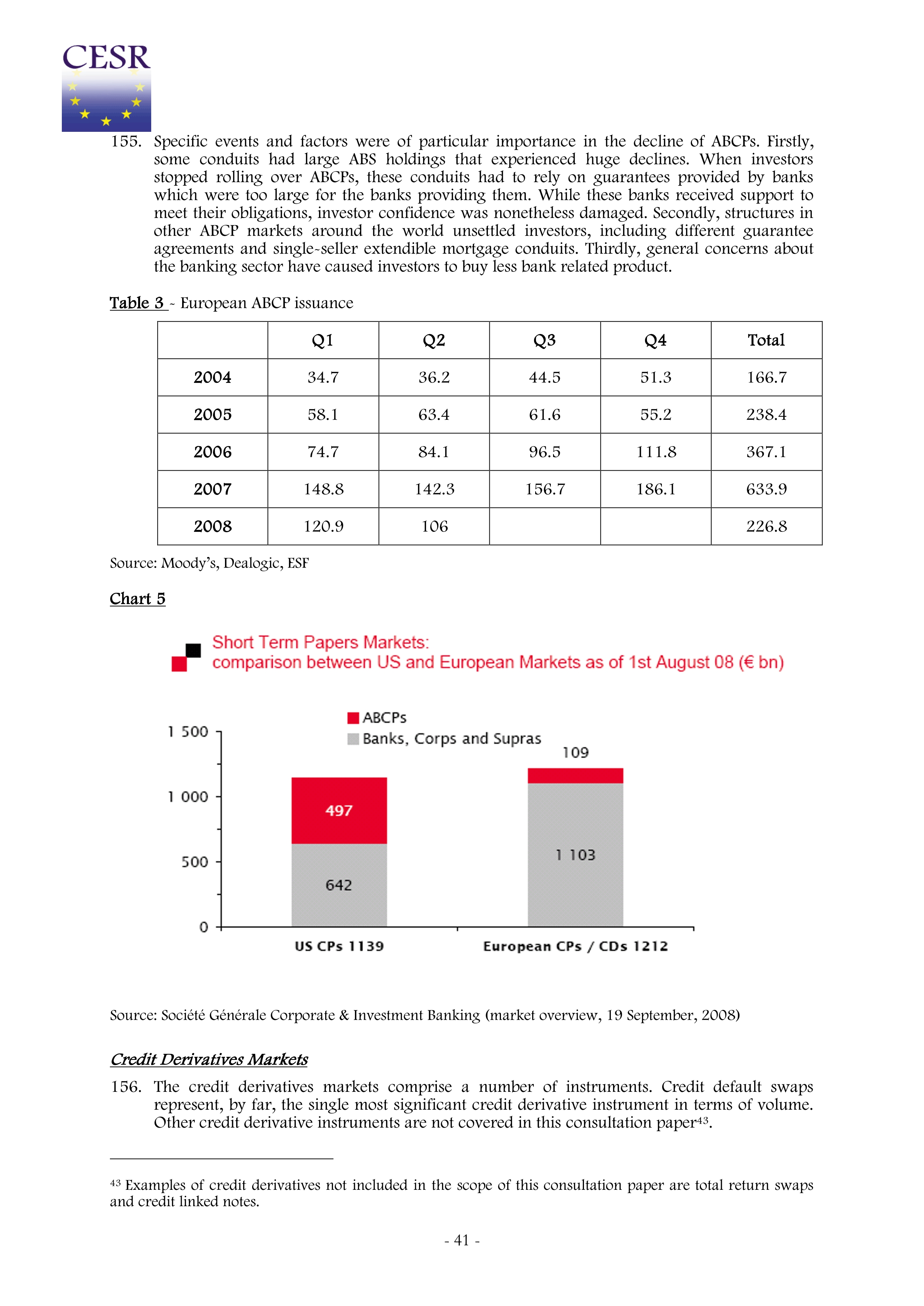}
    \caption{ICDAR 2013}
    \label{fig:icdar13-table}
  \end{subfigure}
  \hfill
  \begin{subfigure}{0.33\linewidth}
    \includegraphics[width=\linewidth]{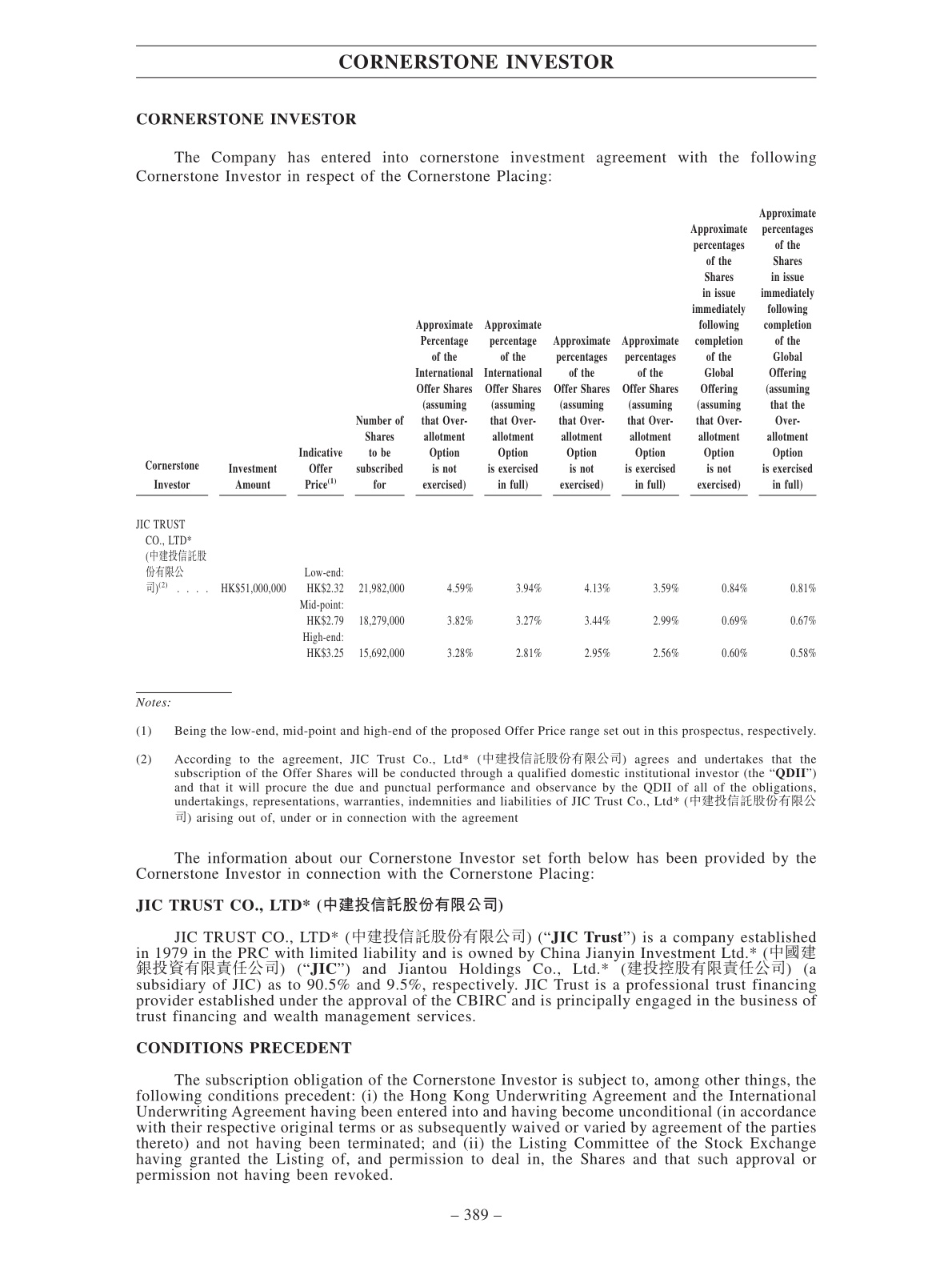}
    \caption{ICDAR 2019}
    \label{fig:icdar19-table}
  \end{subfigure}
  \hfill
  \begin{subfigure}{0.32\linewidth}
    \includegraphics[width=\linewidth]{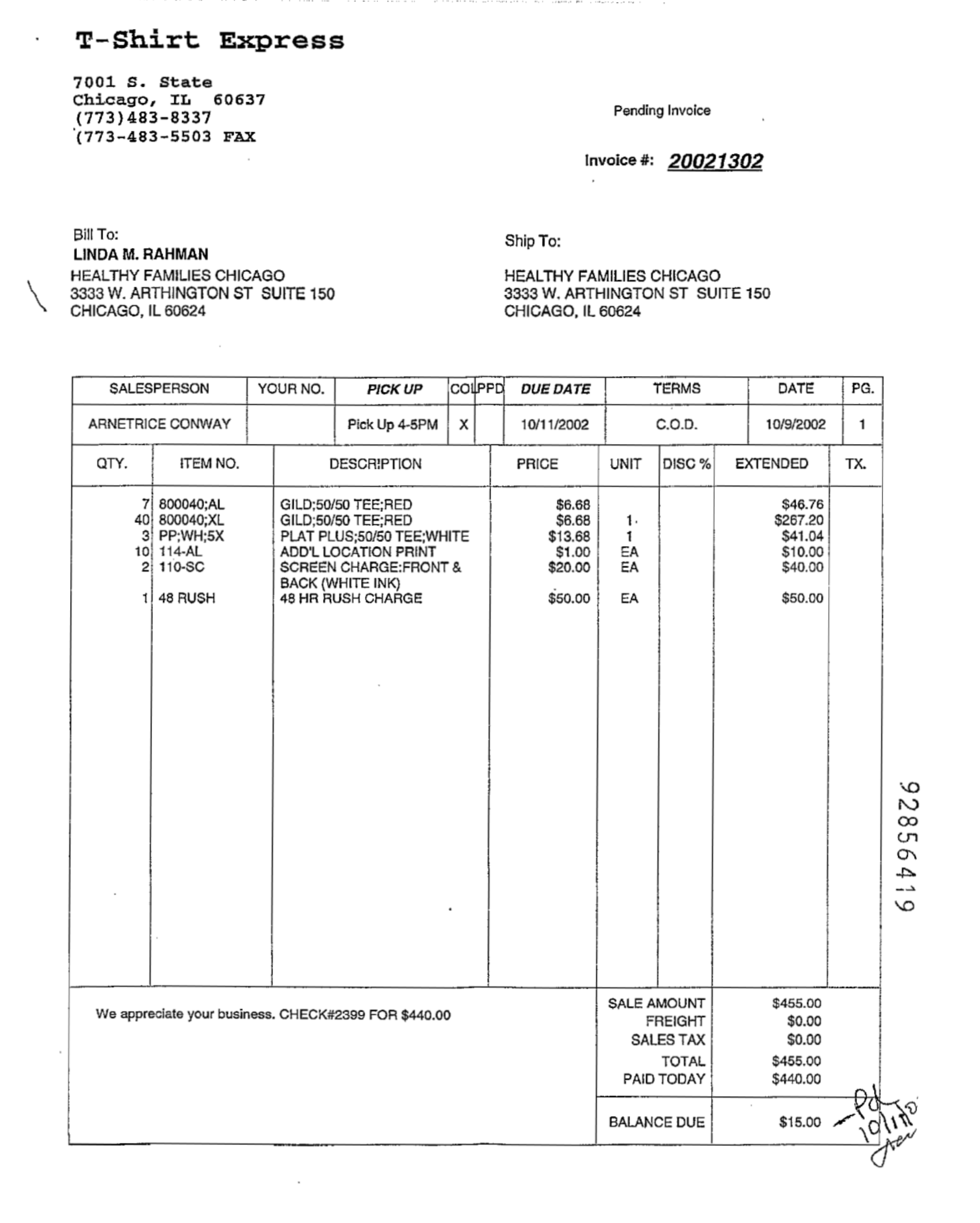}
    \caption{DocILE}
    \label{fig:docile-table}
  \end{subfigure}
  \caption{Examples of documents from the datasets}
  \label{fig:example-tables}
\end{figure*}

We evaluate our approach using public datasets for reproducibility and benchmark comparisons, along with two private datasets to better reflect real-world scenarios. These datasets are described in Table \ref{tab:dataset_details} and Figure \ref{fig:example-tables}.

\begin{table*}[h!]
\centering
\setlength{\tabcolsep}{3pt} 
\begin{tabular}{lcccccc||c}
\hline
\textbf{Dataset}  & \textbf{Acc.} & \textbf{Pages} & \textbf{Cells} & \textbf{Cells/Page} & \textbf{Span. Cells} & \textbf{Type} & \textbf{Use} \\
\hline
\textbf{DocILE}     & Pub.  & 86   & 1085  & 13   & False  & Bus.   & Test   \\
\textbf{IC13}       & Pub.  & 119  & 6301  & 53   & False  & All    & Test   \\
\textbf{IC19}       & Pub.  & 100  & 5074  & 51   & True   & All    & Test   \\
\textbf{Statements} & Priv. & 205  & 9430  & 46   & False  & Prod.  & Test   \\
\textbf{Business}   & Priv. & 208  & 9776  & 47   & False  & Bus.   & Test   \\
\textbf{Refinement}  & Priv. & 50   & 2300  & 46   & False  & Prod.  & Train  \\
\hline
\end{tabular}
\caption{Overview of datasets for RAPTOR training and testing}
\label{tab:dataset_details}

\end{table*}

\subsubsection{Public datasets : ICDAR 2013, ICDAR 2019 and DocILE}

We used the corrected ICDAR 2013, ICDAR 2019, and a subset of the DocILE dataset. Each dataset benchmarks RAPTOR's performance against established standards, ensuring the generalizability and reproducibility of our results.

The ICDAR 2013 dataset \cite{Gbel2013ICDAR2T} is a well-known benchmark for TD and TSR. We utilized the corrected 2023 version available on Hugging Face \cite{Smock2023AligningBD}, which addresses the original dataset's limitations through manual corrections and automated techniques for consistency with other TSR datasets, like PubTables-1M \cite{Smock2021PubTables1MTC}. To better reflect our target documents, which have very few spanning cells, we employed a simplified version of the ICDAR 2013 dataset. We used the competition and non-competition splits from the corrected dataset \cite{Smock2023AligningBD} and performed additional preprocessing by removing all tables with spanning cells and excluding any document pages that no longer contained tables after this filtering.

For a more challenging evaluation, we utilized the ICDAR 2019 dataset \cite{Gao2019ICDAR2C}, specifically the test set for modern tables. We used this dataset unaltered, as its small size and inherent complexity rigorously test RAPTOR's capabilities. The documents contain intricate table structures that differ significantly from the product-oriented tables in our private dataset, making ICDAR 2019 the most demanding benchmark for general table extraction. Additionally, since the original dataset lacks textual annotations, we applied the DOCTR OCR to extract text from the document images.

Lastly, we utilized a subset of the DocILE dataset \cite{vSimsa2023DocILEBF}, which contains business documents like invoices, closely aligning with our private dataset. This makes it a valuable reference for researchers testing similar document types. It’s important to note that the DocILE dataset was not initially designed for tables, so we carefully selected the 86 documents where all columns were annotated, headers were present, and no hierarchical headers or overlapping cells existed. Figures \ref{fig:icdar13-table} and \ref{fig:docile-table} show examples of tables from the corrected ICDAR 2013 and adapted DocILE datasets. These subsets, along with the script used to obtain them, are publicly available here: \href{https://anonymous.4open.science/r/RAPTOR_data-CF2C/README.md}{Data Repository}.

\subsubsection{Private datasets : Business and Statements}

This section outlines the datasets used to evaluate RAPTOR's effectiveness in extracting tables from real-world customer documents. 

The Business dataset includes tables from business files such as invoices and product detail lists. It focuses on simple tables with well-defined structures, avoiding spanning cells, overlapping, or hierarchical headers. A semi-automatic approach was used for its creation: initial table predictions from the baseline pipeline (top of Figure \ref{fig:pipeline}) were manually reviewed and refined by annotators. The entire process (creation, annotation by two annotators, and verification by four), took approximately 20 hours. Figure \ref{fig:example} displays an example of the business dataset. The rest of the document is similar to the DocILE example (Figure \ref{fig:docile-table}).

The Statements dataset was created similarly to maintain consistent annotation and quality, containing tables from statements such as invoice lists.

\subsection{Refinement Dataset}

We used a small Refinement dataset with a structure similar to the Statements dataset. This set was exclusively for parameter tuning using Genetic Algorithms (GA) and for fine-tuning the baseline models in our experiments.

\section{Experiments and results}

RAPTOR's performance is evaluated across various datasets (see Section~\ref{sec:datasets}) against the baseline with all modules disabled. We assess improvements in Table Detection and Table Structure Recognition, conducting an ablation study on the statements dataset to isolate the impact of each refinement. The GA was trained on a subset of 50 documents from the Refinement Dataset, simulating limited training data availability. Results include averages per method with 95\% confidence intervals. Parameter values used are $\alpha = 0.5$, $\beta = 3$, and $\theta_{\text{IoU}}=0.34$.

\subsection{Table Detection}

The baseline we use for comparison is based on TATR and DETR trained on ICDAR 2019, as referenced in Section \ref{sec:related-work}. This baseline operates by taking the predictions from both models, utilizing only their confidence scores to determine the final output. In contrast, we introduce RAPTOR, which enhances these predictions by incorporating refinement modules. To evaluate the performance of both approaches, we employed three metrics: Completeness\cite{Gbel2013ICDAR2T}, Purity\cite{Gbel2013ICDAR2T}, and GIoU \cite{Rezatofighi2019GeneralizedIO}. This structured comparison enables a clear assessment of the effectiveness of our RAPTOR refinements over the baseline system.

Quantitative results in Table \ref{td_results} indicate that our method consistently outperforms or matches the baseline in Purity across all datasets. However, the baseline slightly outperforms our approach in Completeness. While GIoU scores are comparable, the baseline shows a negative average GIoU for the Business dataset, reflecting frequent predictions without overlap with the GT. This evaluation highlights RAPTOR's effectiveness in enhancing TD purity across diverse datasets, although the baseline maintains slightly higher completeness overall.

\begin{table*}[h!]
\centering
\setlength{\tabcolsep}{10pt} 
\renewcommand{\arraystretch}{1.0} 
\begin{tabular}{llc|c} 
\toprule 
\multirow{2}{*}{Dataset} & \multirow{2}{*}{Metric} & \multicolumn{2}{c}{Config} \\ 
\cmidrule(lr){3-4} 
 &  & Baseline \cite{Smock2021PubTables1MTC,Gao2019ICDAR2C} & RAPTOR \\ 
\midrule 
\multirow{3}{*}{DocILE}    & Purity       & 0     & 0     \\  
                           & Completeness & \textbf{58}    & 57   \\  
                           & GIoU         & 9±3     & 9±3     \\  
\hline
\multirow{3}{*}{ICDAR-13}  & Purity       & 0     & 0     \\  
                           & Completeness & \textbf{95}    & 88   \\  
                           & GIoU         & \textbf{18}±1    & 17±2   \\  
\hline
\multirow{3}{*}{ICDAR-19}  & Purity       & 4     & \textbf{9}     \\  
                           & Completeness & \textbf{8}     & 3    \\  
                           & GIoU         & \textbf{16}±2    & 15±2   \\  
\hline
\multirow{3}{*}{Statements} & Purity      & 22    & \textbf{52}    \\  
                           & Completeness & 1     & 1     \\  
                           & GIoU         & 10±2    & \textbf{13}±2    \\  
\hline
\multirow{3}{*}{Business}  & Purity       & 21    & \textbf{35}    \\  
                           & Completeness & \textbf{39}    & 38   \\  
                           & GIoU         & -9±5    & \textbf{5}±3     \\  
\hline
\end{tabular}
\caption{TD results in Purity, Completeness and average GIoU (\%).}

\label{td_results}
\end{table*}

\subsection{Table Structure Recognition}

For the TSR experiments, we utilized TATR as the baseline model, as referenced in the previous section. This task follows the TD step, using the outputs from the TD task as input, creating a more realistic scenario. RAPTOR builds on this by incorporating additional refinement modules (Section \ref{sec:methodology}). We employed Precision, Recall, and F1 metrics from GRITS-CON \cite{Smock2022GriTSGT} to evaluate cell content.

In Table \ref{tsr_detr_detr}, RAPTOR demonstrates significant improvements in business datasets (Statements, Business, and DOCILE), although there is a slight decrease in Recall on the Business dataset. Performance declines on ICDAR-13 and ICDAR-19, particularly in Recall, likely due to the modules reducing the number of predictions by removing too much content.

Overall, RAPTOR meets its objectives by maintaining performance across datasets while enhancing outcomes on targeted product tables. The ablation study on the Statements dataset, shown in Table \ref{ablation_study}, confirms the importance of each refinement. Removing any component results in decreased precision and recall, underscoring their collective significance in RAPTOR’s architecture.

\subsubsection{Impact of RAPTOR Modules on Fine-Tuned Models}

The goal of our modular method is to provide an alternative to fine-tuning when only a limited amount of training data is available. Throughout this work, we assume the availability of only 50 training documents, similar to the Statement dataset. Our comparison in Table \ref{tsr_detr_detr} reveals a limitation of fine-tuning: when applied to such a small dataset, it does not significantly enhance performance. Specifically, while we observe a 15-point increase in precision, there is also a 20-point drop in recall for the \textit{Baseline} model. This table highlights the value of our modular approach, which consistently improves performance across all metrics, even when fine-tuning is employed. Thus, our method not only serves as a viable alternative to fine-tuning but also complements it, providing consistent improvements in model outputs when applied to product tables.

\begin{table*}[h!]
\centering
\setlength{\tabcolsep}{6pt} 
\renewcommand{\arraystretch}{1.0} 
\begin{tabular}{llc|c|c|c}
\toprule
\multirow{2}{*}{Dataset} & \multirow{2}{*}{Metric} & \multicolumn{2}{c}{Original} & \multicolumn{2}{c}{Finetuned} \\
\cmidrule(lr){3-4} \cmidrule(lr){5-6}
 &  & Baseline\cite{Smock2021PubTables1MTC,Gao2019ICDAR2C} & RAPTOR & Baseline\cite{Smock2021PubTables1MTC,Gao2019ICDAR2C} & RAPTOR \\ 
\midrule
\multirow{3}{*}{DocILE} & Precision & 41±6    & \textbf{51}±7        & NA        & NA \\  
                        & Recall & 56±7    & 56±8        &  NA      & NA \\  
                        & F1-score  & 47±8     & \textbf{53}±7        & NA        & NA \\  
\midrule
\multirow{3}{*}{ICDAR-13} & Precision &  \textbf{93}±3   & 92±3        &NA        & NA \\  
                          & Recall & \textbf{84}±4    & 61±7        &NA       & NA \\  
                          & F1-score  & \textbf{88}±3    & 74±4        & NA       & NA \\  
\midrule
\multirow{3}{*}{ICDAR-19} & Precision & 84±4    & \textbf{89}±3        & NA       & NA \\  
                          & Recall & \textbf{61}±6    & 40±6        & NA       & NA \\  
                          & F1-score  & \textbf{71}±5    & 55±4        & NA       & NA \\  
\midrule
\multirow{3}{*}{Statements} & Precision & 61±4    & \textbf{74}±4        & 76±4        & \textbf{85}±3 \\  
                            & Recall & 50±5    & \textbf{64}±5        & 30±4        & \textbf{41}±5 \\  
                            & F1-score  & 55±5    & \textbf{68}±4        & 43±4        & \textbf{55}±4 \\  
\midrule
\multirow{3}{*}{Business} & Precision & 62±5    & \textbf{70}±4        & NA      & NA \\  
                          & Recall & \textbf{54}±5    & 51±5        & NA        & NA \\  
                          & F1-score  & 58±5    & \textbf{59}±5        & NA       & NA \\ 

\bottomrule
\end{tabular}
\caption{TSR results in GRITS-CON (\%).}

\label{tsr_detr_detr}
\end{table*}



\begin{table*}[h!]
\centering
\begin{tabular}{l|c|c|c|c}
\toprule
\textbf{Metric} & \textbf{Baseline} & \textbf{W/O TD refinements} & \textbf{W/O TSR refinements} & \textbf{RAPTOR} \\
\midrule
\textbf{Pre} & 61     &  73   &  71   &    \textbf{74}      \\
\textbf{Rec} & 50     &  60   &  63   &      \textbf{64}   \\
\textbf{F1}  & 55    &  66   &  67   &        \textbf{68}  \\
\bottomrule
\end{tabular}
\caption{Ablation study on the Statements dataset for GRITS-CON (in\%).}
\label{ablation_study}
\end{table*}

\subsection{Computational details}

The average processing time per document is 0.94 ± 0.05 seconds for the baseline system and 1.50 ± 0.15 seconds for the RAPTOR pipeline, adding approximately 0.56 seconds per document. The Genetic Algorithm training utilized parallelism, taking around 23 hours. For fine-tuning, the average training time was about 10 seconds per epoch; the TD models were trained for 250 epochs each, while the TSR model underwent 500 epochs, all on an Nvidia RTX A6000.

\section{Limitations}

RAPTOR achieves strong precision for business documents but struggles with recall on ICDAR-13 and ICDAR-19 datasets\cite{Gbel2013ICDAR2T,Gao2019ICDAR2C}. This stems from its optimization for product tables, which differ from the general tables in these datasets. The noise-removal modules, effective in business contexts, may exclude valid elements in "out-of-scope" tables, reducing recall. Additionally, the system's design assumes single-table inputs by selecting only the highest-scoring bounding box for processing, which limits its application in multi-table documents. The approach's heavy dependence on business document characteristics also constrains its generalizability, requiring significant adjustments for deployment in different domains.

\section{Conclusion}

This paper introduces RAPTOR, a modular system designed to enhance product table extraction from business documents by integrating deep learning for table extraction with specialized modules to address noise and structural errors. By refining parameters using Genetic Algorithms, RAPTOR effectively operates with minimal data. Users can analyze error types and activate relevant modules to improve performance on specific document types.

Evaluations indicate that RAPTOR accurately locates business tables and recognizes their structures while minimizing false positives, which is essential in finance, accounting, report generation, and legal contexts. Additionally, RAPTOR automates table extraction, improving decision-making, productivity, and cost efficiency across various industries. Future work will focus on expanding dataset coverage and enhancing robustness.



\newpage

{\small
\bibliographystyle{splncs04}
\bibliography{egbib}
}

\end{document}